\documentclass{article}

\usepackage{microtype}
\usepackage{graphicx}
\usepackage{subfigure}
\usepackage{booktabs} % for professional tables
\usepackage{multirow} % new
\usepackage{tabularx} % new

\usepackage{hyperref}

\usepackage[accepted]{icml2024}

\usepackage{amsmath}
\usepackage{amssymb}
\usepackage{mathtools}
\usepackage{amsthm}

\usepackage[capitalize,noabbrev]{cleveref}

\theoremstyle{plain}

\theoremstyle{definition}

\theoremstyle{remark}

\usepackage[textsize=tiny]{todonotes}

\icmltitlerunning{Detect Any Text}
\begin{document}

\twocolumn[
\icmltitle{Towards Unified Multi-granularity Text Detection with Interactive Attention}




\begin{icmlauthorlist}
\icmlauthor{Xingyu Wan}{baidu}
\icmlauthor{Chengquan Zhang~\textsuperscript{\rm $\dag$}}{baidu}
\icmlauthor{Pengyuan Lyu}{baidu}
\icmlauthor{Sen Fan}{baidu}
\icmlauthor{Zihan Ni}{baidu}
\icmlauthor{Kun Yao}{baidu}
\icmlauthor{Errui Ding}{baidu}
\icmlauthor{Jingdong Wang}{baidu}
\end{icmlauthorlist}

%\icmlaffiliation{yyy}{Department of XXX, University of YYY, Location, Country}
\icmlaffiliation{baidu}{Baidu, Beijing, China}
%\icmlaffiliation{sch}{School of ZZZ, Institute of WWW, Location, Country}

%\icmlcorrespondingauthor{Firstname1 Lastname1}{first1.last1@xxx.edu}
\icmlcorrespondingauthor{Chengquan Zhang}{zhangchengquan@baidu.com}

\icmlkeywords{Machine Learning, ICML}

\vskip 0.3in
]



\printAffiliationsAndNotice{}  % leave blank if no need to mention equal contribution

\begin{abstract}
Existing OCR engines or document image analysis systems typically rely on training separate models for text detection in varying scenarios and granularities, leading to significant computational complexity and resource demands. 
In this paper, we introduce ``Detect Any Text'' (DAT), an advanced paradigm that seamlessly unifies scene text detection, layout analysis, and document page detection into a cohesive, end-to-end model. 
This design enables DAT to efficiently manage text instances at different granularities, including \emph{word, line, paragraph} and \emph{page}.  
A pivotal innovation in DAT is the across-granularity interactive  attention module, which significantly enhances the representation learning of text instances at varying granularities by correlating structural information across different text queries. 
As a result, it enables the model to achieve mutually beneficial detection performances across multiple text granularities. Additionally, a prompt-based segmentation module refines detection outcomes for texts of arbitrary curvature and complex layouts, thereby improving DAT's accuracy and expanding its real-world applicability.
Experimental results demonstrate that DAT achieves state-of-the-art performances across a variety of text-related benchmarks, including multi-oriented/arbitrarily-shaped scene text detection, document layout analysis and page detection tasks.
\end{abstract}

\section{Introduction}
\label{intro}

%Task explanation:
Text detection serves as the cornerstone for parsing and understanding the content of texts in natural scenes and electronic documents.
Existing document image analysis systems typically categorize text-related detection tasks into separate modules, such as scene text detection, document layout analysis, and document page detection.
Within this context, scene text detection focuses on localizing individual text instances, which may be multi-oriented~\citep{karatzas2015icdar,yao2012detecting} or arbitrarily-shaped~\citep{ch2020total,liu2019curved}, and primarily involves detecting elements at the word-level or text line-level.
Document layout analysis delves into examining geometric structures at the paragraph-level. It involves classifying fine-grained categories within these structures, but does not extend to analyzing their sub-level elements.
Document page detection addresses the identification of the most salient page body in natural scenes, typically utilizing image segmentation techniques~\citep{chen2018encoder,kirillov2023segment}.

\begin{figure}[t]
%\vskip 0.2in
\begin{center}
\centerline{\includegraphics[width=0.98\columnwidth]{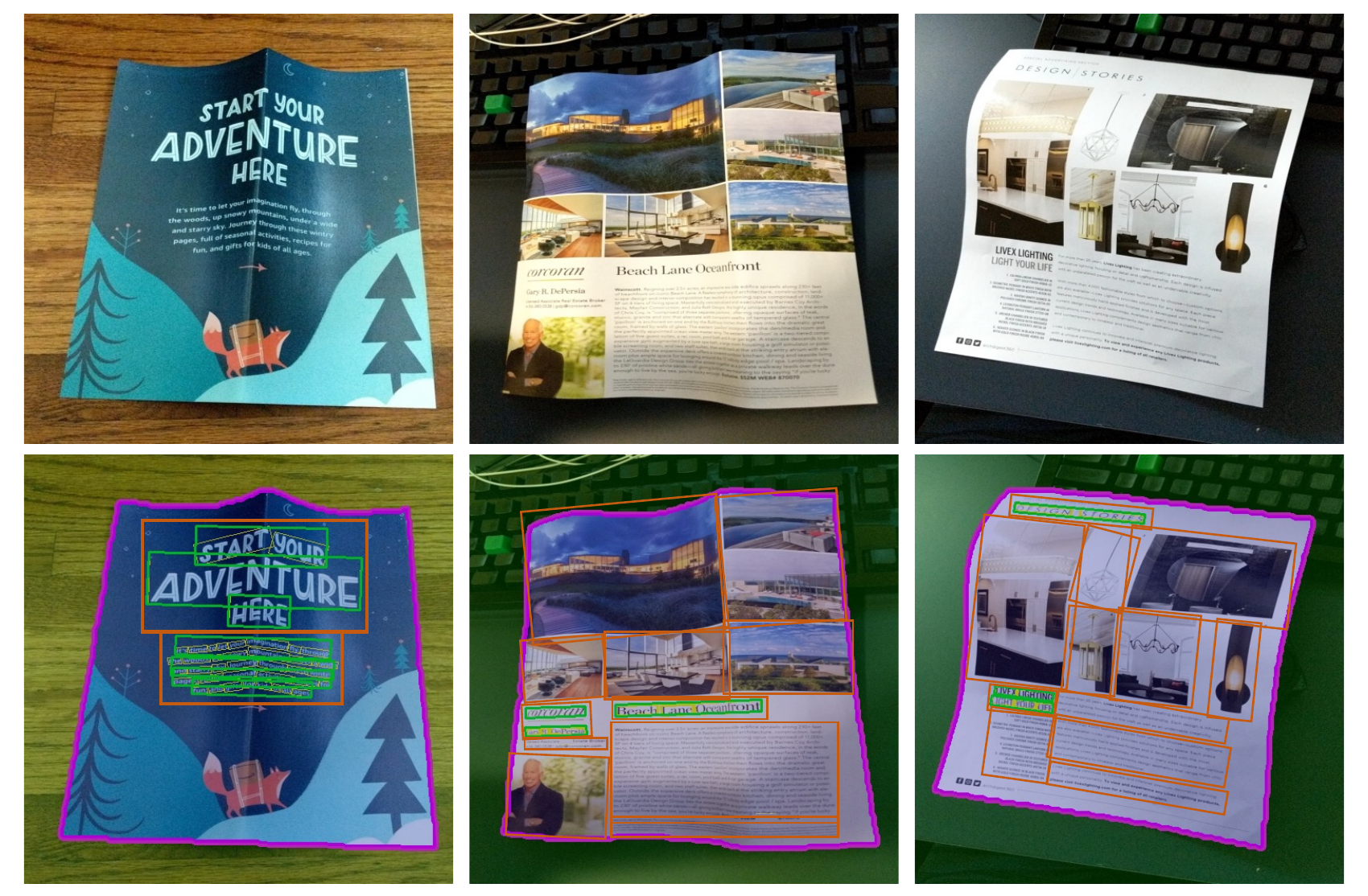}}
\caption{Illustration of the structural correlations among multi-granularity text instances, i.e.,~word(annotated with \textcolor{yellow}{yellow} polygons), text-line(annotated with \textcolor{green}{green} polygons), paragraph(annotated with \textcolor{brown}{brown} polygons) and page(annotated with \textcolor{magenta}{magenta} contours). The blurred small text instances are ignored.}
\label{fig1}
\end{center}
%\vskip -0.3in
\end{figure}

% Problem:
To achieve state-of-the-art (SOTA) results in the aforementioned tasks, current methods necessitate training separate models for each task using diverse datasets, which leads to considerable computational complexity and resource demands.
Moreover, while these tasks involve representation learning at varying text granularities, there is often a lack of attention to the intrinsic correlations among these multi-granularity text instances, as illustrated in~\cref{fig1}.
HierText~\cite{long2022towards} is the first to propose the unified framework for scene text detection and layout analysis, claiming that such a combination can benefit both tasks. However, it has two limitations: (1) It does not fully explore the intrinsic correlations of multi-granularity texts during representation learning. The framework primarily relies on word- or line-based methods for paragraph construction through online clustering, this bottom-up unidirectional approach neglects the potential influence of paragraph-level representations on sub-level elements. (2) The training methodology suffers from limited generalizability due to its reliance on a cascading bottom-up design. This design necessitates ground-truth labeling at all text granularities for each training sample, which restricts its applicability to other prevalent datasets.

%our method:
To overcome these limitations, we introduce DAT, a unified multi-granularity text detection paradigm for detecting text instances at multiple granularities. Unlike the bottom-up, cascaded framework of HierText, DAT incorporates an interactive attention module within its Transformer decoder. This module facilitates the transmission of learned query embeddings across adjacent granularities during representation learning. 
In order to enable parallel training using datasets with incomplete-granularity annotations, we design a multi-granularity detection framework with a mixed-granularity training strategy.
Additionally, to facilitate arbitrarily-shaped text localization and accurate document page segmentation, we follow SAM~\cite{kirillov2023segment} and introduce a prompt-based mask decoder to perform foreground-background segmentation of the multi-granularity text instances.
% contribution:

A key feature of DAT is across-granularity representation learning within the proposed interactive attention module. This module effectively correlates the structural information among text queries of different granularities, enriching the understanding and integration of textual instance representations from both bottom-up and top-down perspectives. This attention mechanism not only elevates the accuracy in text detection but also allows for a more nuanced analysis of texts, regardless of their complexity or format.  This innovative use of interactive attention significantly enhances the versatility and effectiveness of DAT,  making it suitable for a wide range of text detection and understanding scenarios.
% 去掉KNN相关的表述。

%Extensive experiments across various multi-granularity text datasets reveal that:
%(i) By exploring the intrinsic correlation between different text granularities during representation learning,  our interactive attention module substantially improves detection performances across all text granularities, both in top-down and bottom-up manners. %both from words/lines to paragraphs/pages and from paragraphs/pages to words/lines
%(ii) Instead of training separate models for each task, our ``all-in-one'' text detection model, utilizing the mixed-granularity training strategy, leverages text-related datasets with single-granularity annotations for parallel training. This approach addresses the limitation of previous methods that required full annotations at all text levels. Consequently, our model outperforms other SOTA single-task models in benchmarks across scene text detection, document layout analysis, and page segmentation.
%(iii) The introduced mask decoder can perform fine-grained text contour segmentation by using the multi-granularity detection results as prompts, which can significantly improve the detection of arbitrarily-shaped texts, as well as the segmentation of complex layouts and page bodies.

Our contributions can be summarized as follows:
(1) We propose an innovative interactive across-granularity attention module tailored for the representation learning of text instances across varying granularities. 
(2) We design a multi-granularity text detection framework with a mixed-granularity training strategy, which addresses the limitation of previous methods that required full annotations at all text levels. 
%Moreover, it is capable of generating high-quality multi-granularity pseudo labels for incomplete-granularity annotated datasets. 
The resulted model substantially improves detection performances across all text granularities, and outperforms other SOTA single-task models in text detection benchmarks across multiple granularities.
(3) We introduce a prompt-based mask decoder to perform fine-grained text segmentation, which significantly improves the detection performances of arbitrarily-shaped texts, complex layouts and page bodies.

\section{Related Works}
\label{Relat}

\subsection{Text Detection}
%Existing text detection system can be roughly categorized into three tasks, which are scene text detection, document layout analysis and document page detection.

%\subsubsection{Scene Text Detection}
\textbf{Scene Text Detection.} Scene text detection has evolved considerably, primarily divided into two categories: regression-based (or point-based) and segmentation-based approaches.
Regression-based methods~\cite{he2021most,liu2020abcnet,wang2019arbitrary,zhu2021fourier,ye2023dptext} directly regress bounding boxes or polygon points around the text regions, demonstrating their efficiency in detecting texts of varying complexity. To enhance the detection accuracy of texts with arbitrary curvature in complex scenes, the number of regressed points are usually augmented in this line of works.
Segmentation-based methods~\cite{liao2020real, liao2022real,tian2019learning,wang2019efficient,xie2019scene,qin2023towards,wang2020textray,wang2020contournet} frame text detection as a segmentation problem at different levels, e.g.,~pixel level~\cite{liao2020real,liao2022real,tian2019learning,wang2019efficient,xie2019scene,qin2023towards}, segment level~\cite{baek2019character,tang2019seglink++}, and contour level~\cite{wang2020textray,wang2020contournet}, which usually involve grouping algorithms as post-processing stages. This line of works excels at delineating arbitrarily-shaped text by analyzing fine details of contours.
Datasets for scene text detection tasks are typically annotated at the word or text line level granularity.

%\subsubsection{Document Layout Analysis}
\textbf{Document Layout Analysis.} Recent advancements in document layout analysis are marked by the development of comprehensive datasets. 
Notable examples include PubLayNet~\cite{zhong2019publaynet}, DocBank~\cite{li2020docbank}, and DocLayNet~\cite{pfitzmann2022doclaynet}, which offer diverse annotations covering a range of documents from magazines to technical papers. M6Doc~\cite{cheng2023m6doc} is the first dataset to include Chinese examples and blend both real-world and born-digital files, presenting the most fine-grained categories for layout analysis.
Therefore, we use M6Doc to validate the effectiveness of our model in document layout analysis.
The annotation granularity of these datasets is at the paragraph level.

Additionally, HierText~\cite{long2022towards} first proposed a unified framework for scene text detection and layout analysis, claiming that such a combination can simultaneously benefit both tasks. However, due to insufficient representation learning strategy and cascading bottom-up design, its applicability remains confined to specific dataset and scenarios.

%\subsubsection{Document Page Detection}
\textbf{Document Page Detection.} The objective of document page detection (or page frame detection) is to accurately capture the clean and actual contour regions of text page in natural scenes or scanned documents. 
Traditional approaches~\cite{shafait2007page,stamatopoulos2010page,shafait2008document,reza2019robust} primarily utilize detection-based strategies, which involve identifying text regions such as text lines and subsequently employing post-processing techniques to amalgamate these regions into unified page areas.
Modern document image dewarping methods~\cite{das2019dewarpnet,xie2021document,ma2022learning,xue2022fourier} based on deep-learning typically employ image segmentation techniques~\cite{chen2018encoder,kirillov2023segment} to extract the accurate edges of document pages and exclude background information, followed by subsequent rectification processes. This page segmentation is performed on commonly used document dewarping datasets, such as DIW~\cite{ma2022learning} and Doc3D~\cite{das2019dewarpnet}, which are annotated at the page level.

\subsection{Transformer-based Object Detection}
Recent advancements in text detection have been significantly influenced by the evolution of Transformer-based object detection algorithms. 
A pivotal development in this field was marked by DETR~\cite{carion2020end}, which introduced a novel one-to-one label assignment strategy and eliminated the need for manually designed components like non-maximum suppression (NMS).
Subsequent methods have delved deeper into the evolution of decoder queries within DETR~\cite{carion2020end}. Deformable DETR~\cite{zhu2021deformable} proprosed a deformable attention module that focuses on specific sampling points around a reference point. DN-DETR~\cite{li2022dn} introduced a denoising training method by bringing noisy annotations and boxes to the decoder. DINO~\cite{zhang2022dino} advanced this field further by introducing mixed query selection and contrastive denoising modules. Most recently, Group-DETR~\cite{chen2023group} proposed to learn group-wise object queries for one-to-many label assignment, enhancing both detection accuracy and training efficiency.
% DETR first. More variants towards deeper understanding of decoder quries: DINO, groupDETR.

Different from Group-DETR~\cite{chen2023group}, our proposed DAT model adopts multiple groups of object queries to enable parallel training for multi-granularity text detection, and to facilitate the correlation of intrinsic structural information across different text granularities.
In the DAT decoder, each group of object queries is distinctly defined by text instances at varying granularities, including \emph{word, line, paragraph} and \emph{page}.

\section{Method}
% conclude of each section
\begin{figure*}[t]
%\vskip 0.2in
\begin{center}
\centerline{\includegraphics[width=0.98\textwidth]{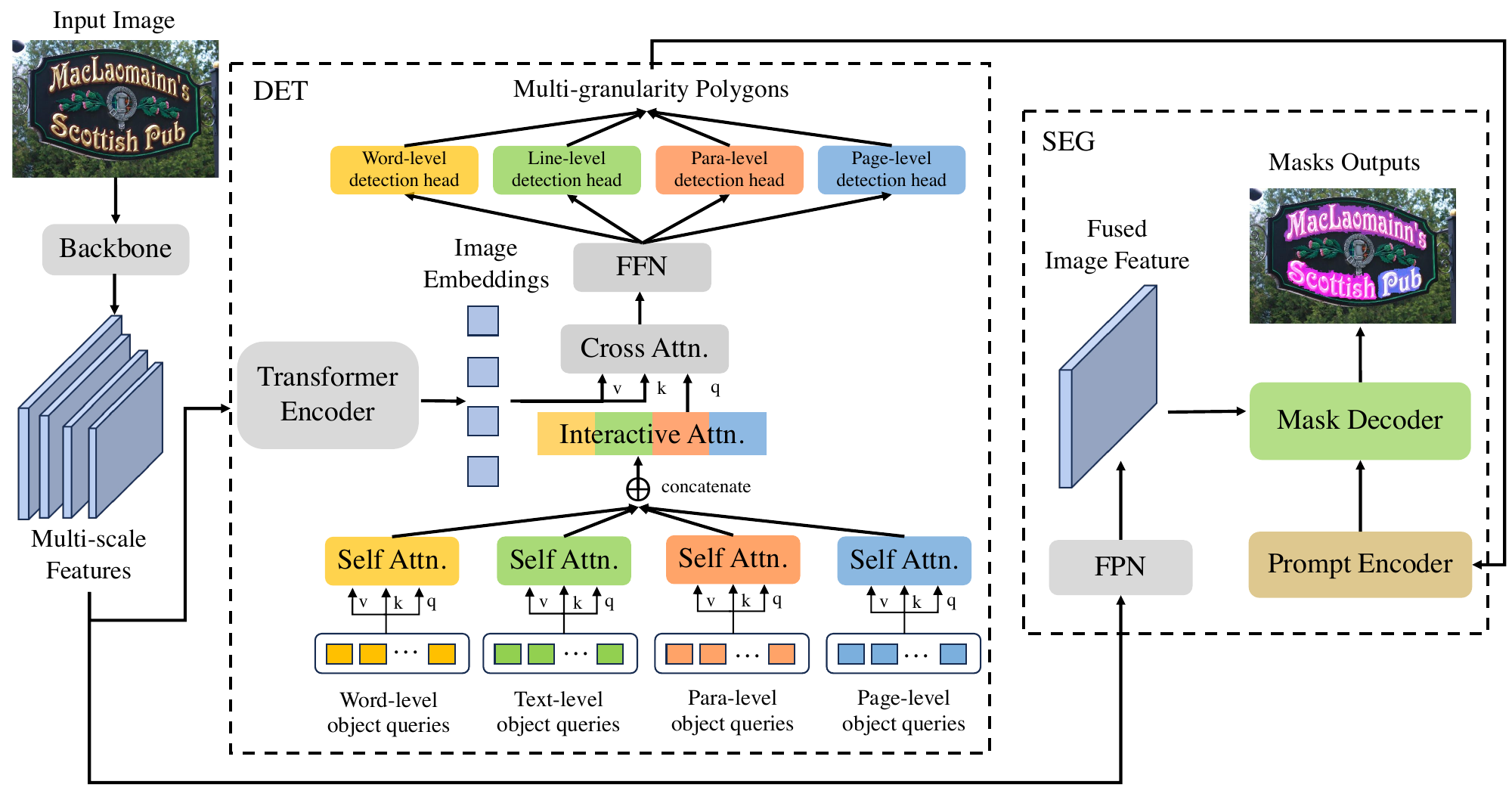}}
\caption{Network structure of Detect Any Text (DAT). ``DET'' illustrates the multi-granularity detection framework with a single layer of Transformer decoder network, where the residual connection and norm layers are omitted for simplicity. ``SEG'' illustrates the model pipeline of prompt-based segmentation module.}
\label{fig2}
\end{center}
%\vskip -0.3in
\end{figure*}

\subsection{Preliminaries}
\label{subsec1}
We frame multi-granularity text detection task as two hierarchical branches, i.e., text detection (DET) and text segmentation (SEG), as demonstrated in~\cref{fig2}.
Given an input image $\textbf{I} \in \mathbb{R}^{3 \times H \times W}$, the output of DET branch is defined as $\textbf{Y}^{DET}=\{(b_{j},c_{j})\}_{j=1}^{n} $, where $b_{j} \in \mathbb{R}^{4}$ denotes the polygon points coordinates of $j$-th located text instance, and $c_{j} \in \mathcal{C}^{DET}$ denotes the corresponding assigned class label. Here, the category vocabulary $\mathcal{C}^{DET}=\mathcal{C}^{word}+\mathcal{C}^{line}+\mathcal{C}^{para}+\mathcal{C}^{page}$ is composed of four granularity levels. Specifically, $\mathcal{C}^{para}$ denotes a multi-class vocabulary, while $\mathcal{C}^{word}$, $\mathcal{C}^{line}$, and $\mathcal{C}^{page}$ represent binary classifications.
The output of SEG branch is defined as $\textbf{Y}^{SEG}=\{m_{j}\}_{j=1}^{n}$, where $m_{j} \in \mathbb{R}^{1 \times H \times W}$ denotes the predicted mask of $j$-th detected text instance.
The forward propagation for DET ans SEG branches are formulated as follows:
\begin{equation}
\label{eq1}
\textbf{Y}^{DET} = f_{\mathcal{H}}(f_{dec}(f_{enc}(\textbf{F})|\textbf{A}, \textbf{Q}))
\end{equation}
\begin{equation}
\label{eq2}
%Y^{SEG} = f_{\mathcal{M}}(f_{enc}(f_{fpn}(\textbf{F})), Y^{DET})
\textbf{Y}^{SEG} = f_{\mathcal{M}}(f_{fpn}(\textbf{F}), \textbf{Y}^{DET})
\end{equation}
For DET branch in Eq.\eqref{eq1}, the Transformer encoder $f_{enc}(\cdot)$ first aggregates the multi-scale image features $\textbf{F}$ using multi-head self-attention, and the Transformer decoder $f_{dec}(\cdot)$ takes the aggregated image embedding, attention mask $\textbf{A}$ and group queries $\textbf{Q}$ as inputs to conduct interactive feature learning and global reasoning about text instances. The output $\textbf{Y}^{DET}$ is obtained through a multi-task detection head $f_{\mathcal{H}}(\cdot)$ for each granularity.
For SEG branch in Eq.\eqref{eq2}, we adopt a FPN network $f_{fpn}(\cdot)$ to obtain a fused image feature from $\textbf{F}$. The task-agnostic mask decoder $f_{\mathcal{M}(\cdot)}$ takes the fused image feature and detection output $\textbf{Y}^{DET}$ as inputs to conduct detection-conditioned image segmentation.

\subsection{Multi-granularity Detection Framework}
\label{subsec3.2}
We employ the advanced Transformer-based object detection algorithm DINO~\cite{zhang2022dino} to construct our text detection framework. To enable parallel training and inference for multi-granularity text instances, we initialize a set of learnable query embeddings for each granularity of text instance separately, forming group queries $\textbf{Q}=\{(\textbf{Q}_{k}^{word}, \textbf{Q}_{k}^{line}, \textbf{Q}_{k}^{para}, \textbf{Q}_{k}^{page})\}_{k=1}^{N_{q}}$ that serve as inputs to the Transformer decoder as in Eq.\eqref{eq1}. Here $N_{q}$ is the query number of each group, which is same for all text granularities.
As in~\cref{fig2}, each layer of Transformer decoder is composed of three components: 1) group-wise self-attention module with non-shared parameters for learning text queries at each granularity, 2) an interactive across-granularity attention module for correlating the intrinsic structural information between different text queries (introduced in \labelcref{subsec3}), 3) a parameter-shared cross-attention module and feed-forward network (FFN) for global reasoning about text instances at each text granularity. Additionally, a group-wise multi-task detection head $f_{\mathcal{H}}=\{f_{\mathcal{H}}^{word},f_{\mathcal{H}}^{line},f_{\mathcal{H}}^{para},f_{\mathcal{H}}^{page}\}$  is added to the FFN. Here $f_{\mathcal{H}}$ for each granularity is composed of a box regression head, a box classification head and a polygon regression head, in which the polygon regression head is only added to the last layer of Transformer decoder for efficiency.

\textbf{Mixed-granularity Training.} The training target for optimizing multi-granularity text detection task is defined as follow:
\begin{equation}
\label{eq3}
\mathcal{L}^{DET} = \sum_{t=1}^{4}( \omega_t \times \mathcal{L}_{t}(\textbf{Y}_{t}^{DET}, \hat{\textbf{Y}}_{t}^{DET}))
\end{equation}
The subscript $t$ in Eq.\eqref{eq3} refers to four different tasks of optimizing \emph{word, line, paragraph, page} granularities. $\textbf{Y}_{t}^{DET}$ is the output prediction for granularity $t$, and $\hat{\textbf{Y}}_{t}^{DET}$ is the corresponding training label generated from ground-truth(GT). The loss function of each granularity is composed of $l_{1}$ loss for polygon regression, $l_{1}$ and GIoU~\cite{rezatofighi2019generalized} losses for box regression, and focal loss~\cite{lin2017focal} for classification. The loss weights $\omega_{t}$ for multi-task learning are defined as follow:
\begin{align}
\label{eq4}
\omega_{t} =
    \begin{cases}
      0, & \text{if } \hat{N}_{t}=0; \\
      1, & \text{if } \hat{N}_{t}=1 \text{ and } \sum_{t=1}^{4}\hat{N}_{t}=1; \\
      \frac{1}{\sum_{t=1}^{4}\hat{N}_{t}}, & \text{if } \hat{N}_{t}=1 \text{ and } \sum_{t=1}^{4}\hat{N}_{t}>1.
    \end{cases}
\end{align}

Here $\hat{N}_{t}$ is a binary indicator $(0/1)$ representing whether the label of granularity $t$ is annotated in the GT. It is worth mentioning that the loss weights $\omega_{t}$ for each text granularity are dynamically adjusted within each training batch.

\textbf{Discussion.} Such framework design leverages the power of parallel training on diverse datasets, even those with limited annotation granularities such as single-granularity annotations or incomplete labeling schemes. Notably, the resulting model generates multi-granularity text detection outputs in one-forward-propagation, leading to significantly improved efficiency in text-related systems. 
Moreover, our model is capable of generating high-quality multi-granularity pseudo labels for incomplete-granularity annotated datasets. The detailed results of generated pseudo labels are present in Sec~\labelcref{visualize}.

\subsection{Across-granularity Representation Learning with Interactive Attention Module}
\begin{figure}[t]
%\vskip 0.2in
\begin{center}
\centerline{\includegraphics[width=0.98\columnwidth]{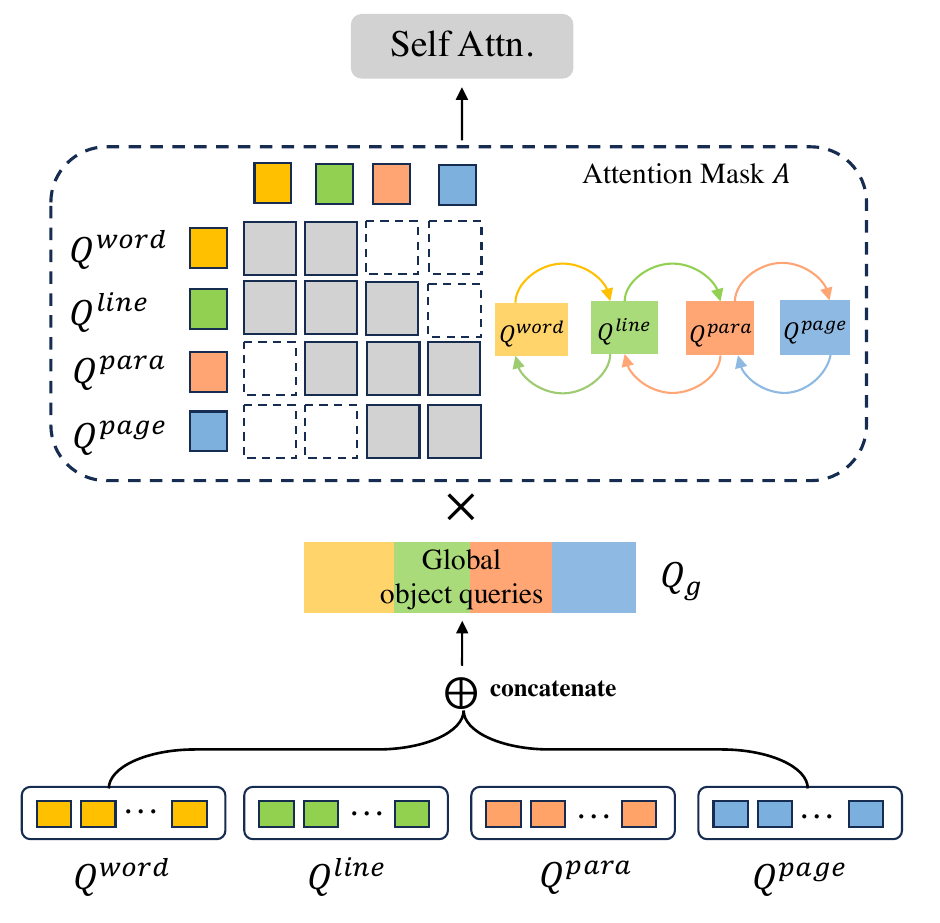}}
\caption{Illustration of across-granularity representation learning with interactive attention module (interaction factor $\mathcal{I}=1$).}
\label{fig3}
\end{center}
%\vskip -0.3in
\end{figure}

%\subsection{Interactive Across-granularity Attention Module}
\label{subsec3}
%\textbf{Motivation.} 
%Existing deep-learning based OCR engines or document image analysis systems inevitably involve representation learning at different level of text instances to handle any granularity of text detection tasks. 
As shown in~\cref{fig1}, text instances in natural scenes or document images are normally (but not necessarily) correlated to each other structurally among different granularities. Most existing approaches overlooked the correlation of these intrinsic linked multi-level texts, while we argue that such intrinsic correlations can be useful to facilitate a deeper understanding and integration of textual instance representations. 
%\textbf{Across-granularity interactive attention.} 
Motivated by this, we introduce an across-granularity interactive attention module to text detection decoder, facilitating the transmission of learned query embeddings across adjacent granularities during representation learning. As shown in~\cref{fig3}, after group-wise self-attention for each level of query embeddings, we concatenate them to form a global query embedding $\textbf{Q}_{g} \in \mathbb{R}^{4N_{q} \times c}$, here $c$ is the embedding dimensions for each query. We employ a global attention mask $\textbf{A}$ with interaction factor $\mathcal{I}$ to conduct across-granularity self-attention for global query embedding $\textbf{Q}_{g}$. The attention mask $\textbf{A}$ is a binary matrix with a shape of $4N_{q} \times 4N_{q}$ (here we omit the batch size and numbers of attention heads for simplicity).
In this layer of global self-attention, the interactions between query embeddings of different granularities depend on the weight parameters at corresponding positions in the attention mask $\textbf{A}$. 
When $\mathcal{I}=1$, the global query embedding is enabled to interactive across different levels of query embeddings only in adjacent granularities, i.e., the interactions of word-to-line, line-to-para, para-to-page from bottom-up, and page-to-para, para-to-line, line-to-word from top-down. When $\mathcal{I}$ is increased to 2 and 3, more extensive cross-granularity interactions are allowed during global self-attention. %Specifically, the bi-directional interactions between word-para, line-page are enabled when $k=2$, and word-page is enabled when $k=3$. It is worth mentioning that when $k=3$, query embeddings of different granularities are fully connected, meaning no mask is applied to this global self-attention module.
%We observed in experimental results that $\mathcal{I}$ brings the best performance for most of the text detection datasets.
After this interactive across-granularity attention computation, we extract each group of query embeddings from the corresponding positions of global query embedding $\textbf{Q}_{g}$ to obtain the updated $\textbf{Q}=\{(\textbf{Q}_{k}^{word}, \textbf{Q}_{k}^{line}, \textbf{Q}_{k}^{para}, \textbf{Q}_{k}^{page})\}_{k=1}^{N_{q}}$ for the subsequent cross-attention.

%The proposed across-granularity attention module can effectively correlate the intrinsic structural information among text queries by learning representations from other granularity of query embeddings and enabling the duplicate removal for instances. By doing so, it facilitates a deeper understanding and integration of textual instance representations from bottom-up and top-down, ranging from individual words to entire page.

\subsection{Prompt-based Segmentation Module}
To address the problem of arbitrarily-shaped text localization and accurate document page segmentation, we introduce a hierarchical prompt-based segmentation module to perform foreground-background segmentation of the multi-granularity text instances. As illustrated in~\cref{fig2}, the learnable parameters for segmentation module is composed of a FPN layer for extracting a fused image feature, a Prompt Encoder for representing multi-granularity polygons from detection module, and a Mask Decoder for generating fine-grained masks of given text regions. Following SAM~\cite{kirillov2023segment}, we initialize a group of learnable embeddings to sum with the positional encodings of each polygon coordinates for encoding multi-granularity polygons within Prompt Encoder. The Mask Decoder includes blocks of prompt self-attention, two-way cross-attention, up-sampling, and MLP modules.

The introduced mask decoder can perform more fine-grained text contour segmentation by using the multi-granularity detection results as prompts, which can significantly improve the detection of curved and arbitrarily-shaped texts, as well as the segmentation of complex layouts and page bodies.

The training target for optimizing segmentation module is a linear combination of mean-square-error(MSE) and dice losses~\cite{milletari2016v} for mask prediction and MSE loss for intersection-over-union(IoU) prediction:
\begin{align}
\label{eq5}
\mathcal{L}^{SEG} = & \; 5 \times \mathcal{L}_{mse}(\textbf{Y}^{SEG}, \hat{\textbf{Y}}^{SEG}) \nonumber \\
& + \mathcal{L}_{dice}(\textbf{Y}^{SEG}, \hat{\textbf{Y}}^{SEG}) \nonumber \\
& + \mathcal{L}_{iou}(\textbf{Y}^{SEG}, \hat{\textbf{Y}}^{SEG})
\end{align}

\begin{table*}[ht]
\caption{Results for DAT and other SOTA models on benchmark test sets of scene text detection, layout analysis, and document page segmentation. ``WORD, LINE, PARA, PAGE'' indicate the word detection, text-line detection, layout analysis and page segmentation respectively. ``P, R, F'' stand for Precision, Recall, Fscore metrics. The best and secone-best metrics are highlighted in \textbf{bold} and \color{blue}{blue}.}
\label{tab_benchmark}
%\vskip 0.15in
\begin{center}
%\begin{small}
\begin{scriptsize}
\setlength{\tabcolsep}{2.5pt} % 减小列间距
\begin{sc}
\begin{tabular}{c|ccc|ccc|ccc|ccc|c|c}
\toprule
\multirow{4}{*}{method} & \multicolumn{6}{c|
}{word} & \multicolumn{6}{c|}{line} & \multicolumn{1}{c|}{para} & \multicolumn{1}{c}{page} \\
\cmidrule(r){2-7} \cmidrule(r){8-13} \cmidrule(r){14-14} \cmidrule(r){15-15}
& \multicolumn{3}{c|}{ICDAR2015} & \multicolumn{3}{c|}{Total-Text} & \multicolumn{3}{c|}{CTW1500} & \multicolumn{3}{c|}{MSRA-TD500} & \multicolumn{1}{c|}{M6Doc} & \multicolumn{1}{c}{DIW} \\
\cmidrule(r){2-4} \cmidrule(r){5-7} \cmidrule(r){8-10} \cmidrule(r){11-13} \cmidrule(r){14-14} \cmidrule(r){15-15}
& P & R & F & P & R & F & P & R & F & P & R & F & mAP & mIoU \\
\midrule
SAM~\cite{kirillov2023segment} & - & - & - & - & - & - & - & - & - & - & - & - & - & 84.1  \\
DeepLabv3+~\cite{chen2018encoder} & - & - & - & - & - & - & - & - & - & - & - & - & - & \color{blue}{98.61} \\
M6Doc~\cite{cheng2023m6doc} & - & - & - & - & - & - & - & - & - & - & - & - & \color{blue}{63.8} & -  \\
HierText~\cite{long2022towards} & - & - & - & 85.49 & \textbf{90.53} & 87.94 & 84.56 & 87.44 & 85.97 & 88.04 & \color{blue}{87.44} & 87.70  & - & -  \\
SIR~\cite{qin2023towards} & 90.4 & 85.4 & 87.8 & 90.9 & 85.6 & 88.2 & 87.4 & 83.7 & 85.5 & \color{blue}{93.6} & 86.0 & 89.6 & - & -  \\
DPText-DETR~\cite{ye2023dptext} & - & - & - & 91.8 & 86.4 & 89.0 & \color{blue}{91.7} & 86.2 & 88.8 & - & - & - & - & -  \\
UNITS~\cite{kil2023towards} & \textbf{94.0} & 91.0 & \color{blue}{92.5} & - & - & 89.8 & - & - & - & - & - & - & - & -  \\
ESTextSpotter~\cite{huang2023estextspotter} & \color{blue}{92.5} & 89.6 & 91.0 & 92.0 & 88.1 & 90.0 & 91.5 & 88.6 & \color{blue}{90.0} & 92.9 & 86.3 & 89.5 & - & -  \\
%TCM~\cite{yu2023turning} & - & - & 89.2 & - & - & - & - & - & 84.9 & - & - & 88.8 & - & -  \\
\midrule
DAT-DET (ours) & 90.87 & \color{blue}{94.51} & \textbf{92.66} & \color{blue}{93.98} & 88.17 & \color{blue}{90.98} & 89.25 & \color{blue}{89.28} & 89.26 & \textbf{95.11} & 86.63 & \textbf{90.67} & - & -  \\
DAT-SEG (ours) & 87.46 & \textbf{95.76} & 91.42 & \textbf{95.04} & \color{blue}{89.16} & \textbf{92.01} & \textbf{92.51} & \textbf{90.94} & \textbf{91.72} & 92.74 & \textbf{88.60} & \color{blue}{90.62} & \textbf{65.7} & \textbf{98.65}  \\

\bottomrule
\end{tabular}
\end{sc}
%\end{small}
\end{scriptsize}
\end{center}
%\vskip -0.2in
\end{table*}

\section{Experiments}
\subsection{Experiment Setup}
\label{setup}

\textbf{Datasets and Evaluation Protocol.}
Our experimental framework utilized popular benchmarks corresponding to each level of text granularity. For word detection, we used the ICDAR2015~\cite{karatzas2015icdar} and Total-Text~\cite{ch2020total} datasets; for line detection, CTW1500~\cite{liu2019curved} and MSRA-TD500~\cite{yao2012detecting} were employed; M6Doc~\cite{cheng2023m6doc} facilitated our document layout analysis; and DIW~\cite{ma2022learning} was the choice for page detection.
Notably, ICDAR2015 and MSRA-TD500 are multi-oriented datasets annotated with quadrilateral points, Total-Text and CTW-1500 feature arbitrarily shaped texts annotated with polygon points outlining text contours. M6Doc offers a fine-grained layout analysis with 74 categories, and DIW is recognized for document dewarping, annotated with foreground page masks.
For evaluation metrics, we report Precision, Recall, and F1-Score (abbreviated as ``P, R, F") for word and line detection tasks. The mean Average Precision (mAP) metric is used for layout analysis, and mean Intersection Over Union (mIoU) is used for page segmentation tasks.

\textbf{Implementation Details.}
We adopted the Swin Transformer Large (SwinL)~\cite{liu2021swin} pretrained on ImageNet-22K~\cite{deng2009imagenet} as our initialization backbone network. For comprehensive benchmark evaluations, we trained our DAT model on a diverse set of datasets: ICDAR2015~\cite{karatzas2015icdar}, Total-text~\cite{ch2020total}, Curved SynthText~\cite{liu2020abcnet}, ICDAR-MLT~\cite{nayef2017icdar2017}, ArT~\cite{chng2019icdar2019}, CTW1500~\cite{liu2019curved}, MSRA-TD500~\cite{yao2012detecting}, M6Doc~\cite{cheng2023m6doc}, DIW~\cite{ma2022learning}, and Doc3d~\cite{das2019dewarpnet}. Notably, ICDAR-MLT and ArT are multilingual datasets, with annotations for Chinese and Japanese texts at the text-line level, and annotations for other languages at the word level. To accommodate these datasets within our DAT framework, we implemented a masking strategy. Specifically, for images annotated with Chinese or Japanese texts, we masked out texts of other languages, categorizing these images under the text-line level for training. Conversely, images devoid of these languages were categorized under the word level. 
%This strategy allowed for the effective incorporation of multilingual datasets while preserving the granularity integrity in our training. 
Further elaboration on training settings is detailed in the Appendix.

\begin{table*}[ht]
\caption{Ablation study on impact of each text granularity. ``WORD, LINE, PARA, PAGE'' indicate the word detection, text-line detection, layout analysis and page detection respectively. ``P, R, F'' stand for Precision, Recall, Fscore metrics.}
\label{tab_ab1}
%\vskip 0.1in
\begin{center}
%\begin{small}
\begin{scriptsize}
\setlength{\tabcolsep}{2.5pt} % 减小列间距
\begin{sc}
\begin{tabular}{c|ccc|ccc|ccc|ccc|c|c}
\toprule
\multirow{4}{*}{model} & \multicolumn{6}{c|
}{word} & \multicolumn{6}{c|}{line} & \multicolumn{1}{c|}{para} & \multicolumn{1}{c}{page} \\
\cmidrule(r){2-7} \cmidrule(r){8-13} \cmidrule(r){14-14} \cmidrule(r){15-15}
& \multicolumn{3}{c|}{ICDAR 15} & \multicolumn{3}{c|}{Total-Text} & \multicolumn{3}{c|}{CTW1500} & \multicolumn{3}{c|}{MSRA-TD500} & \multicolumn{1}{c|}{M6Doc} & \multicolumn{1}{c}{DIW} \\
\cmidrule(r){2-4} \cmidrule(r){5-7} \cmidrule(r){8-10} \cmidrule(r){11-13} \cmidrule(r){14-14} \cmidrule(r){15-15}
& P & R & F & P & R & F & P & R & F & P & R & F & mAP & mIoU \\
\midrule
single granularity baseline & 72.90 & 91.19 & 81.02 & 89.82 & 89.66 & 89.74 & 81.05 & 80.02 & 80.53 & 91.46 & 83.87 & 87.50 & 62.0 & \textbf{98.67}  \\
word + line & 82.37 & \textbf{97.16} & 89.15 & 88.29 & \textbf{90.97} & 89.61 & 85.73 & 80.48 & 83.02 & 92.78 & 84.95 & 88.69 & - & - \\
word + line + para & 88.99 & 94.17 & 91.51 & 91.56 & 90.20 & 90.88 & 88.90 & 88.92 & 88.91 & 94.94 & 86.48 & 90.51 & 67.8 & -  \\
word + line + para + page (DAT) & \textbf{90.87} & 94.51 & \textbf{92.66} & \textbf{93.98} & 88.17 & \textbf{90.98} & \textbf{89.25} & \textbf{89.28} & \textbf{89.26} & \textbf{95.11} & \textbf{86.63} & \textbf{90.67} & \textbf{70.5} & 98.65  \\

\bottomrule
\end{tabular}
\end{sc}
%\end{small}
\end{scriptsize}
\end{center}
%\vskip -0.25in
\end{table*}

\begin{table*}[ht]
\caption{Analysis of different attention modules. The best results are highlighted in \textbf{bold}.}
\label{tab_ab2}
%\vskip 0.1in
\begin{center}
%\begin{small}
\begin{scriptsize}
\setlength{\tabcolsep}{2.5pt} % 减小列间距
\begin{sc}
\begin{tabular}{c|ccc|ccc|ccc|ccc|c|c}
\toprule
\multirow{4}{*}{model} & \multicolumn{6}{c|
}{word} & \multicolumn{6}{c|}{line} & \multicolumn{1}{c|}{para} & \multicolumn{1}{c}{page} \\
\cmidrule(r){2-7} \cmidrule(r){8-13} \cmidrule(r){14-14} \cmidrule(r){15-15}
& \multicolumn{3}{c|}{ICDAR 15} & \multicolumn{3}{c|}{Total-Text} & \multicolumn{3}{c|}{CTW1500} & \multicolumn{3}{c|}{MSRA-TD500} & \multicolumn{1}{c|}{M6Doc} & \multicolumn{1}{c}{DIW} \\
\cmidrule(r){2-4} \cmidrule(r){5-7} \cmidrule(r){8-10} \cmidrule(r){11-13} \cmidrule(r){14-14} \cmidrule(r){15-15}
& P & R & F & P & R & F & P & R & F & P & R & F & mAP & F@0.9 \\
\midrule
\textit{w/o} interactive attention  & 76.32 & 87.67 & 81.60 & 85.80 & 83.79 & 84.78 & 77.63 & 78.39 & 78.01 & 92.23 & 82.03 & 86.83 & 54.9 & 85.31  \\
bottom-up attention & 81.56 & 95.62 & 88.03 & 90.44 & 90.20 & 90.32 & 83.60 & 83.21 & 83.40 & 88.77 & \textbf{89.86} & 89.31 & 63.0 & 89.05 \\
interactive attention ($\mathcal{I}=1$) & \textbf{90.88} & 94.51 & \textbf{92.66} & \textbf{93.98} & 88.17 & 90.98 & \textbf{89.25} & \textbf{89.28} & \textbf{89.26} & \textbf{95.11} & 86.63 & \textbf{90.67} & 70.5 & \textbf{94.15}  \\
interactive attention ($\mathcal{I}=2$) & 83.47 & \textbf{97.01} & 89.73 & 90.76 & \textbf{91.37} & \textbf{91.06} & 86.48 & 86.11 & 86.30 & 90.28 & \textbf{89.86} & 90.07 & \textbf{71.2} & 91.00  \\
interactive attention ($\mathcal{I}=3$) & 76.99 & 89.74 & 82.88 & 86.69 & 84.42 & 85.54 & 75.66 & 79.01 & 77.30 & 89.57 & 84.48 & 86.96 & 61.2 & 87.25 \\

\bottomrule
\end{tabular}
\end{sc}
%\end{small}
\end{scriptsize}
\end{center}
%\vskip -0.2in
\end{table*}

\subsection{Main Results}
\label{benchmark}
%We evaluated our proposed DAT model and compared it with SOTA methods across various text granularities. The results are summarized in~\cref{tab_benchmark}.
As outlined in Section~\labelcref{subsec3}, the DAT model consists of two key branches: a text detection branch (DAT-DET) and a hierarchical segmentation branch (DAT-SEG). \cref{tab_benchmark} demonstrates that our ``all-in-one'' DAT model consistently outperforms single-task models, achieving SOTA performances in all text-related tasks: scene text detection, document layout analysis, and page segmentation.

\textbf{Scene Text Detection.}
The DAT-DET model outperformed the previous SOTA methods UNITS~\cite{kil2023towards} and SIR~\cite{qin2023towards} on ICDAR2015 and MSRA-TD500 multi-oriented datasets. The DAT-SEG model further boosted the F-score by 2.01 and 1.72 points on Total-Text and CTW-1500 datasets espectively, outperforming ESTextSpotter~\cite{huang2023estextspotter}.
Our approach obtained the highest precision on arbitrarily shaped datasets and the highest recall on multi-oriented datasets, validating the effectiveness of our multi-granularity text detection framework. Notably, the DAT-SEG model showed a slight decrease in performance on ICDAR2015 and MSRA-TD500 datasets compared to the DAT-DET model. This is attributed to the quadrilateral-based annotations used in these datasets, which does not favor the more refined outcomes of DAT-SEG. We discuss this further with visualizations in Sec~\labelcref{visualize}.

\textbf{Document Layout Analysis.}
For paragraph-level document layout analysis, the DAT-SEG model significantly improved performance on M6Doc dataset~\cite{cheng2023m6doc}, currently the most fine-grained dataset with 74 categories. The model's mAP saw an increase from 63.8 to 65.7, a notable gain of 1.9 points. This improvement indicates that our model, leveraging interactive information from multiple granularities, is adept at learning more discriminative and nuanced paragraph features, enhancing its overall document layout analysis capability.

\textbf{Document Page Detection.}
We implemented SAM~\cite{kirillov2023segment} on DIW dataset with official pre-trained weights, employing its default Automatic Mask Generation configuration. The largest mask area from panoptic segmentation results was chosen as the final page segmentation outcome. In contrast, DeepLabv3+~\cite{chen2018encoder} was fine-tuned on DIW and Doc3d training sets for a fair comparison. As shown in the last column of~\cref{tab_benchmark}, our DAT-SEG model outperformed these algorithms on DIW dataset. This success is attributed to our robust representation learning for individual text elements and accurate page segmentation via edge features. 

%The last column of~\cref{tab_benchmark} illustrates that our DAT-SEG model achieves competitive, and in some cases superior, results on the DIW dataset when compared to these SOTA algorithms. This performance advantage is partially attributable to our model's robust representation learning phase for text instances, where an interactive attention module extracts nuanced sub-level feature information. Additionally, our prompt-based segmentation module effectively captures the fine-grained edge features of the document pages' main bodies, thereby significantly enhancing the overall segmentation accuracy.

%\textbf{Qualitative Results.}
% visually improved on ICDAR15 and MSRA-TD500

\begin{figure*}[t]
%\vskip 0.2in
\begin{center}
\centerline{\includegraphics[width=0.96\textwidth]{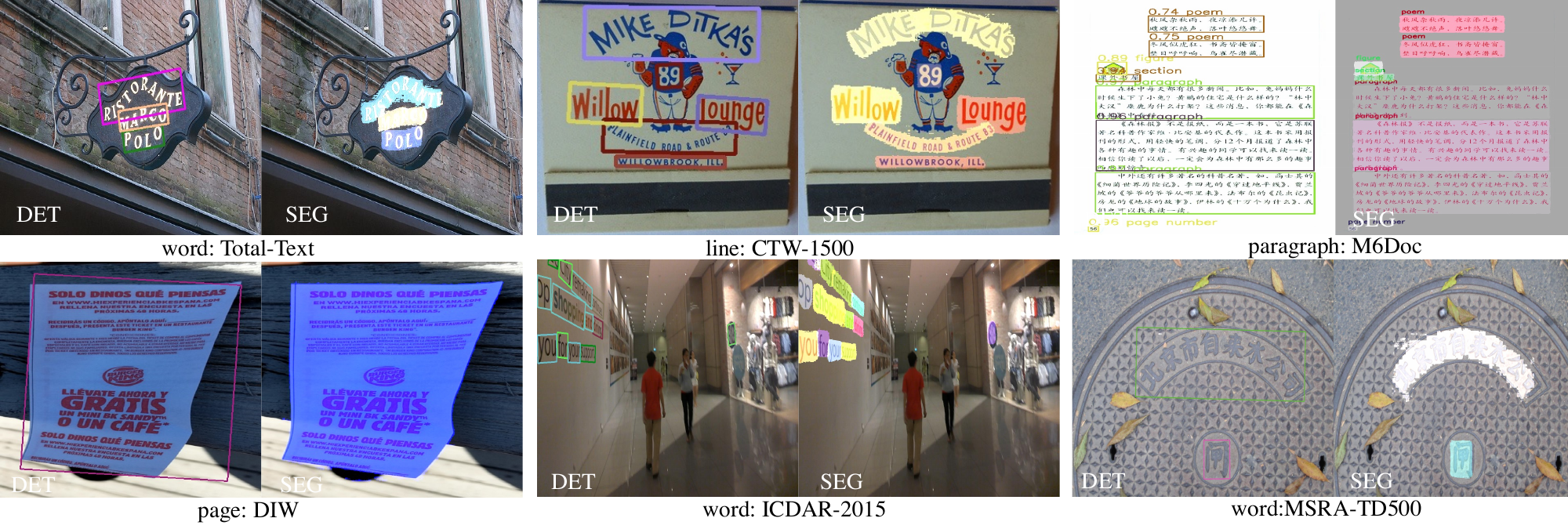}}
\caption{Visualization results of DAT on each granularity of benchmark datasets. ``DET'' and ``SEG'' indicate text detection and segmentation results respectively. For multi-oriented datasets ICDAR-2015 and MSRA-TD500, the DAT-SEG model further refined detection results, particularly for curved texts. However, a slight decline in benchmark evaluation results occurred due to the quadrilateral-based annotations.}
\label{fig4}
\end{center}
%\vskip -0.3in
\end{figure*}

\begin{figure}[t]
%\vskip 0.2in
\begin{center}
\centerline{\includegraphics[width=0.98\columnwidth]{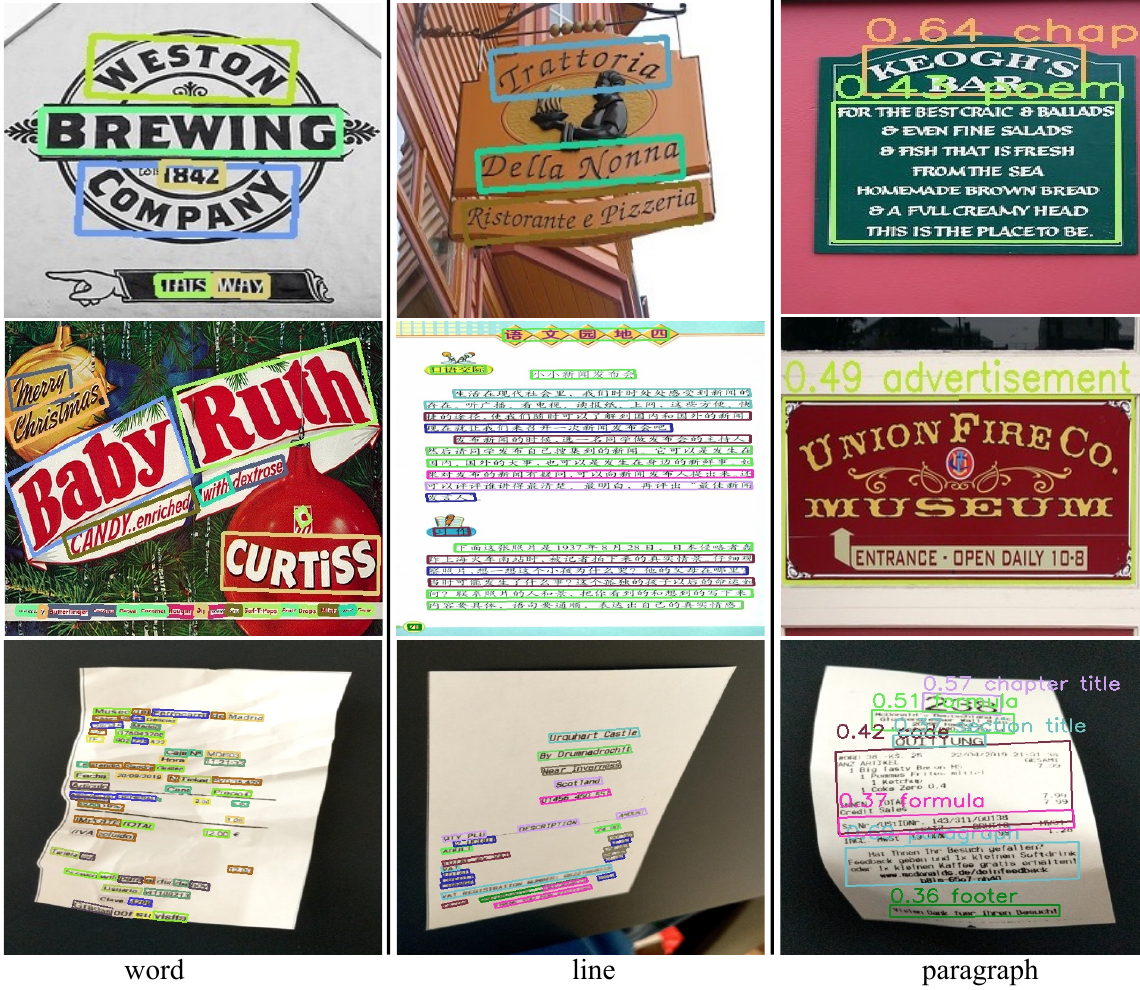}}
\caption{Multi-granularity pseudo labels produced by DAT. From left to right: text detection results at the word, line and paragraph levels. Note that these datasets do not have the corresponding GT annotations for these specific granularities.}
\label{fig5}
\end{center}
\vskip -0.2in
\end{figure}

\subsection{Ablation Study}
\label{ablation}

\subsubsection{Impact of each text granularity}

\textbf{Word + Line.} 
\cref{tab_ab1} reveals that the $word+line$ model saw an 8.13-point F-score increase at the word-level on ICDAR2015 dataset, compared to the $baseline$ model (\cref{tab_ab1} row 1). However, on Total-Text dataset, a minor decrease of 0.13-point in F-score was observed. This drop is primarily attributed to a rise in false positives due to the integration of line-level features, as indicated by a higher recall but lower precision. On CTW1500 and MSRA-TD500 datasets, the model registered F-score improvements of 2.49 and 1.19 points respectively, showcasing the efficacy of word-level features in supporting line-level detection.

\textbf{Word + Line + Paragraph.}
As detailed in~\cref{tab_ab1}, incorporating paragraph granularity tasks led to significant enhancements across various granularity benchmarks. Notably, the $word+line+para$ model showed a remarkable 5.89-point increase in F-score on CTW1500 dataset over the $word+line$ model. Furthermore, on M6Doc dataset, this model exhibited an impressive mAP improvement from 62.0 to 67.8 (+5.8), confirming our hypothesis that text line distributions are beneficial for detailed layout analysis and that layout structures can guide the localizations of words and text lines.

\textbf{Word + Line + Paragraph + Page.}
Our full DAT model (\cref{tab_ab1} row 4), which includes page-level granularity, achieved the highest performance metrics in text detection tasks across word, line, and paragraph granularities. This highlights the value of page-level granularity in providing top-down guidance for sub-level text detection. On DIW dataset, the full model experienced a slight mIoU decrease from 98.67 to 98.65 compared to the $baseline$ model, due to the relatively simpler task of page segmentation. Nevertheless, the $baseline$ model's robust performance on this dataset already outperformed DeepLabv3+~\cite{chen2018encoder} (98.61). The introduction of page detection in our full model leverages top-down feature learning at the page level, significantly enhancing sub-level text detection tasks.

\subsubsection{Analysis of interactive attention module}

% Loss curve. -> convergence speed
\textbf{Without Interactive Attention.}
The first row of~\cref{tab_ab2} presents the model's performance without the interactive attention module. This model relies solely on group-wise self-attention and global cross-attention within the Transformer decoder. The evaluation results show that, while the model was trained using datasets of various granularities, it only attained sub-optimal detection outcomes. This underlines the importance of interactive attention in enhancing the model’s learning capabilities for better performance.

\textbf{Bottom-up Attention.}
Similar to HierText~\cite{long2022towards}, we constructed a bottom-up attention scheme to investigate the effects of unidirectional interactive attention. As shown in the second row of~\cref{tab_ab2}, incorporating bottom-up attention significantly improved text detection across all granularities compared to models without interactive attention. Notably, this approach improved the F-score on Total-Text from 87.94 to 90.32 (+2.38), and on MSRA-TD500 from 87.70 to 89.31 (+1.61), surpassing original HierText metrics. Unlike HierText, our method does not require hierarchical text annotations and benefits single granularity detection by increasing training data volume.

\textbf{Interactive Attention.}
We observed the highest detection performance metrics at the line and page granularities when $\mathcal{I}=1$ as demonstrated in ~\cref{tab_ab2}. Increasing $\mathcal{I}$ to 2 improved performance on Total-Text and M6Doc but resulted in a decline in other datasets such as ICDAR2015, CTW1500, and DIW, where F-scores dropped by 2.93, 2.96, and 3.15, respectively. This change indicates a trade-off between higher recall and lower precision, suggesting that $\mathcal{I}=2$ model introduced more false positives, reducing text detection accuracy. When $\mathcal{I}$ was further increased to 3, there was a sharp drop in performance across all granularities, nearly mirroring the model without interactive attention. This implies that a completely unrestricted information interaction (no attention mask) is detrimental to feature learning across granularities due to difficulty in distinguishing relevant features from noise. Therefore, we selected $\mathcal{I}=1$ for our experimental benchmark evaluations on public datasets.

\subsection{Qualitative Results}
\label{visualize}
% visually improved on ICDAR15 and MSRA-TD500
% Figure 4. Some qualitative detection examples using our model with SOTA methods.
\cref{fig4} presents the qualitative results of DAT across different text granularities. 
Our model shows notable detection performances in arbitrarily-shaped datasets Total-Text and CTW-1500. The model also demonstrates its proficiency in fine-grained paragraph classification on M6Doc dataset, as well as accurate page segmentation on DIW dataset.
For the multi-oriented datasets ICDAR-2015 and MSRA-TD500 (the last two blocks in~\cref{fig4}, the segmentation results output by the DAT model after introducing the prompt-based segmentation module further optimize the text contours within the detected polygons, especially for curved texts. This demonstrates the effectiveness of the proposed segmentation module for the overall multi-granularity detection framework.
Thanks to the multi-granularity detection framework design and the across-granularity interactive attention module, our DAT model is capable of generating high-quality pseudo labels for incomplete-granularity annotated datasets as demonstrated in~\cref{fig5}, more detailed analyses are provided in the Appendix.

\subsection{Discussion}
\label{exp-discussion}
\textbf{Computational Cost Analysis.}
The number of parameters (Params) and GFLOPS of our proposed DAT-DET model are 228.29M and 394 respectively, and the complete DAT model (DET+SEG) has a Params of 284.65M and GFLOPS of 474. 
%We also tested the results of other single-granularity SOTA methods on the same A100 NVIDIA GPU. For instance, the Params and GFLOPS of DPText-DETR~\cite{ye2023dptext} are 67.82M and 249.
Our method has approximately 2 times the GFLOPS of single-task SOTA method DPText-DETR~\cite{ye2023dptext} (GFLOPS=249), but the training/testing speeds remain competitive or even faster than SOTA methods. For instance, the training and testing FPS of our DAT-DET model are 1.4 and 3.57, respectively. In contrast, the training and testing FPS of DPText-DETR~\cite{ye2023dptext} are 0.56 and 3.84, respectively.
Moreover, our approach achieves the detection of text at four different granularities using a single unified model, with the only cost being a slight increase in GPU memory usage (31.21G). In contrast, previous SOTA methods dedicated to single-task operations would require training and testing separate models for multi-granularity tasks, essentially necessitating four times the amount of time for training and testing.
These analyses not only highlight our model's superior efficiency and effectiveness but also underscore its innovation in handling multi-granularity text detection tasks within a single framework.

\textbf{Limitations.}
The limitations of our DAT method can be summarized as follows:
(1) The number of parameters and GFLOPS of our DAT model are relatively larger than previous single-task text detection models, requiring more GPU memory and a longer training cycle, but the unified framework is still more cost-effective than the sum of the three independent task models, as discussed above. (2) Our model shows a low utilization rate of multilingual training data (such as MLT and ArT). Due to the different annotation granularities of different languages in existing multilingual datasets (e.g., English annotations at the word level, Chinese at the line level), we have not yet been able to clarify whether the annotation granularity of other languages is word or line, except for Chinese and English. As a result, we have only used incomplete training samples when training the MLT set (described in~Section~\labelcref{setup}). In the future, we will explore the possibility of unified training that includes more languages with different granularities.

\section{Conclusion}
In this paper, we introduce a novel multi-granularity text detection paradigm, termed as ``DAT''. Inspired by the inherent structural relationships among different text granularities in natural scenes, we propose a bi-directional interactive attention module within the text detection decoder to bolster representation learning across all granularities. Particularly, our approach is distinguished by its independence from complete granularity data annotations and the capability of parallel training of a single model for concurrent text detection tasks across word, line, paragraph, and page granularities within a unified detection framework. Our extensive experiments on public datasets reveal that this module significantly enhances text detection performance at all levels of granularity, establishing a new benchmark for State-of-the-Art (SOTA) in multi-granularity text detection models. Additionally, we have integrated a prompt-based segmentation module to accurately localize arbitrarily-shaped texts and segment document pages. These innovative designs allow our model to outperform other SOTA single-task models across a variety of benchmarks, including scene text detection, document layout analysis, and page segmentation. 
%Our method shows great potential for broad application in a variety of text detection and comprehension tasks.

\section*{Impact Statement}
This paper presents work whose goal is to advance the field of Machine Learning. There are many potential societal consequences of our work, none which we feel must be specifically highlighted here.


\bibliography{DAT}
\bibliographystyle{icml2024}

%%%%%%%%%%%%%%%%%%%%%%%%%%%%%%%%%%%%%%%%%%%%%%%%%%%%%%%%%%%%%%%%%%%%%%%%%%%%%%%
%%%%%%%%%%%%%%%%%%%%%%%%%%%%%%%%%%%%%%%%%%%%%%%%%%%%%%%%%%%%%%%%%%%%%%%%%%%%%%%

% APPENDIX
%%%%%%%%%%%%%%%%%%%%%%%%%%%%%%%%%%%%%%%%%%%%%%%%%%%%%%%%%%%%%%%%%%%%%%%%%%%%%%%
%%%%%%%%%%%%%%%%%%%%%%%%%%%%%%%%%%%%%%%%%%%%%%%%%%%%%%%%%%%%%%%%%%%%%%%%%%%%%%%
\newpage
\appendix
\onecolumn

\section{Training Details}
Our multi-granularity text detection framework was implemented on 8 NVIDIA A100 GPUs. During the model training phase, the batch size is set to 8 (1 per single GPU). The query number $N_{q}$ of each group (Sec~\labelcref{subsec3.2}) is set to 900. We train our full DAT model using public datasets of all granularities, with total 120 epochs. The base learning rate is $1 \times 10^{-4}$ and reduced to $1 \times 10^{-5}$ at the 66-th epoch and $1 \times 10^{-6}$ at the 99-th epoch. For our proposed mixed-granularity training(Sec~~\labelcref{subsec3.2}), the weight of $l_{1}$ loss is $5.0$, the weight of GIoU loss is $2.0$, and the weight of focal loss is $1.0$. We choose AdamW with a weight decay parameter of $1 \times 10^{-4}$ as our optimizer. The number of both encoder and decoder layers is set to $6$. The size of fused image feature after the FPN layer (~\cref{fig2}) is $\frac{1}{8}$ of the original input image size. We adopt multiple data augmentation strategies for training DAT-DET module including: 1) randomly flipping the image with a probability of 0.3; 2) randomly rotating the image within a range of -45 to 45 degrees with a probability of 0.5; 3) random color distortion with a probability of 0.1; 4) randomly cropping the image with a probability of 0.3; 5) randomly resizing the shorter size of input images within a range of 480 to 800 with an interval of 32, while constraining the longer size within 1333 pixels. No additional data augmentation strategies were used when training the DAT-SEG module.
During the model testing phase, we uniformly resize the input images to $800 \times 1333$ in height and width.

\begin{figure*}[ht]
%\vskip 0.2in
\begin{center}
\centerline{\includegraphics[width=\textwidth]{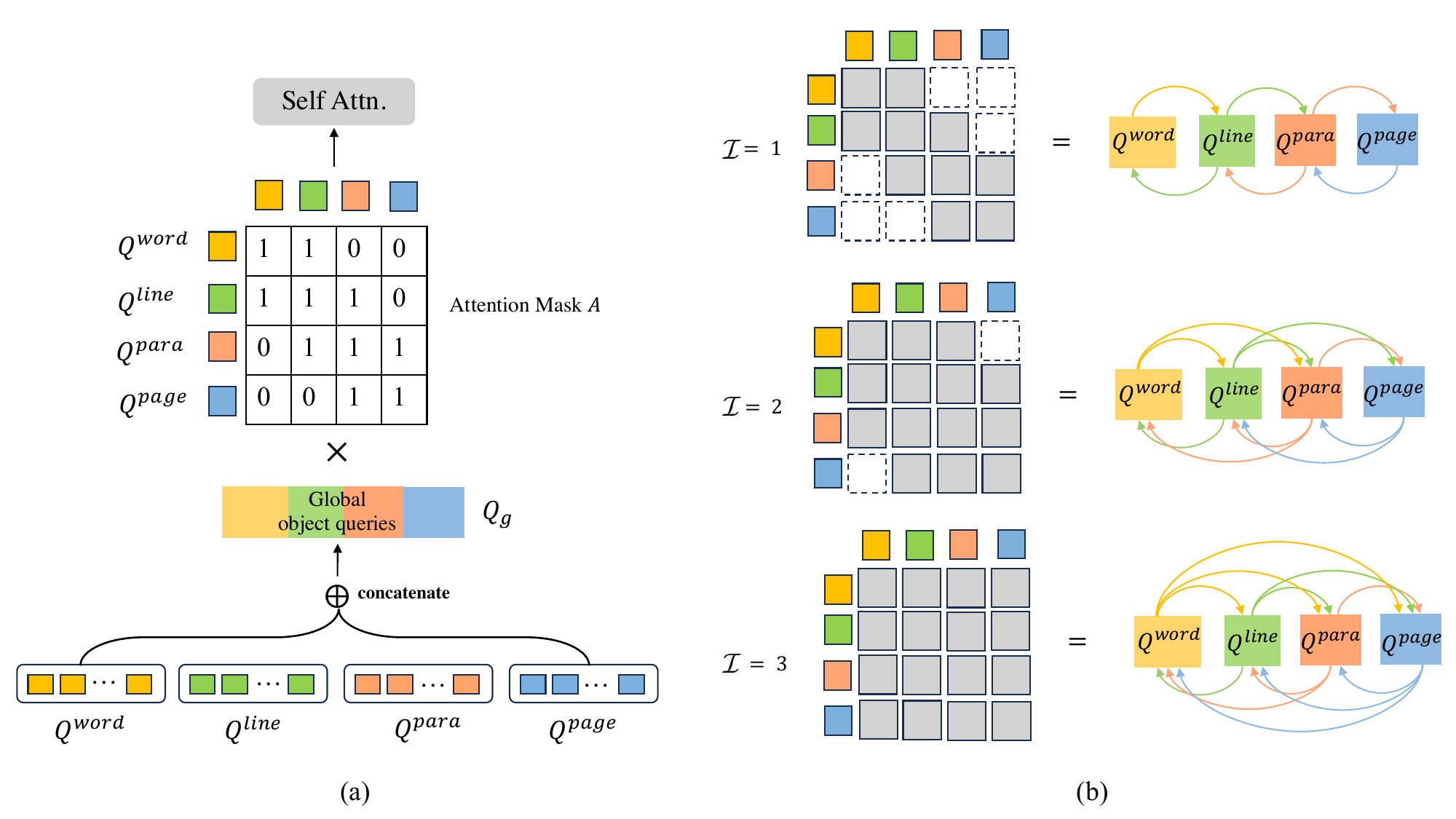}}
\caption{Illustration of interactive attention module with different interaction factors $\mathcal{I}$.}
\label{app_fig1}
\end{center}
%\vskip -0.2in
\end{figure*}

\section{Explanation for the interactive attention module with different interaction factors}
Within our proposed multi-granularity text detection framework, the feature interactions between text queries of different granularities are facilitated through a global self-attention layer, which is guided by an attention mask $\textbf{A}$. This interactive can be adjusted based on the interactive factor $\mathcal{I}$. Specifacally, as illustrated in~\cref{app_fig1},
when $\mathcal{I}=1$, the global query embedding is enabled to interactive across different levels of query embeddings only in adjacent granularities, i.e., the interactions of word-to-line, line-to-para, para-to-page from bottom-up, and page-to-para, para-to-line, line-to-word from top-down. When $\mathcal{I}$ is increased to 2 and 3, more extensive cross-granularity interactions are allowed during global self-attention. Specifically, the bi-directional interactions between word-para, line-page are enabled when $\mathcal{I}=2$, and word-page is enabled when $\mathcal{I}=3$. It is worth mentioning that when $\mathcal{I}=3$, query embeddings of different granularities are fully connected, meaning no mask is applied to this global self-attention module.

The proposed across-granularity attention module can effectively correlate the intrinsic structural information among text queries by learning representations from other granularity of query embeddings and enabling the duplicate removal for instances. By doing so, it facilitates a deeper understanding and integration of textual instance representations from bottom-up and top-down, ranging from individual words to entire page.

\section{Qualitative results of multi-granularity pseudo labels}
\begin{figure*}[ht]
%\vskip 0.2in
\begin{center}
\centerline{\includegraphics[width=\textwidth]{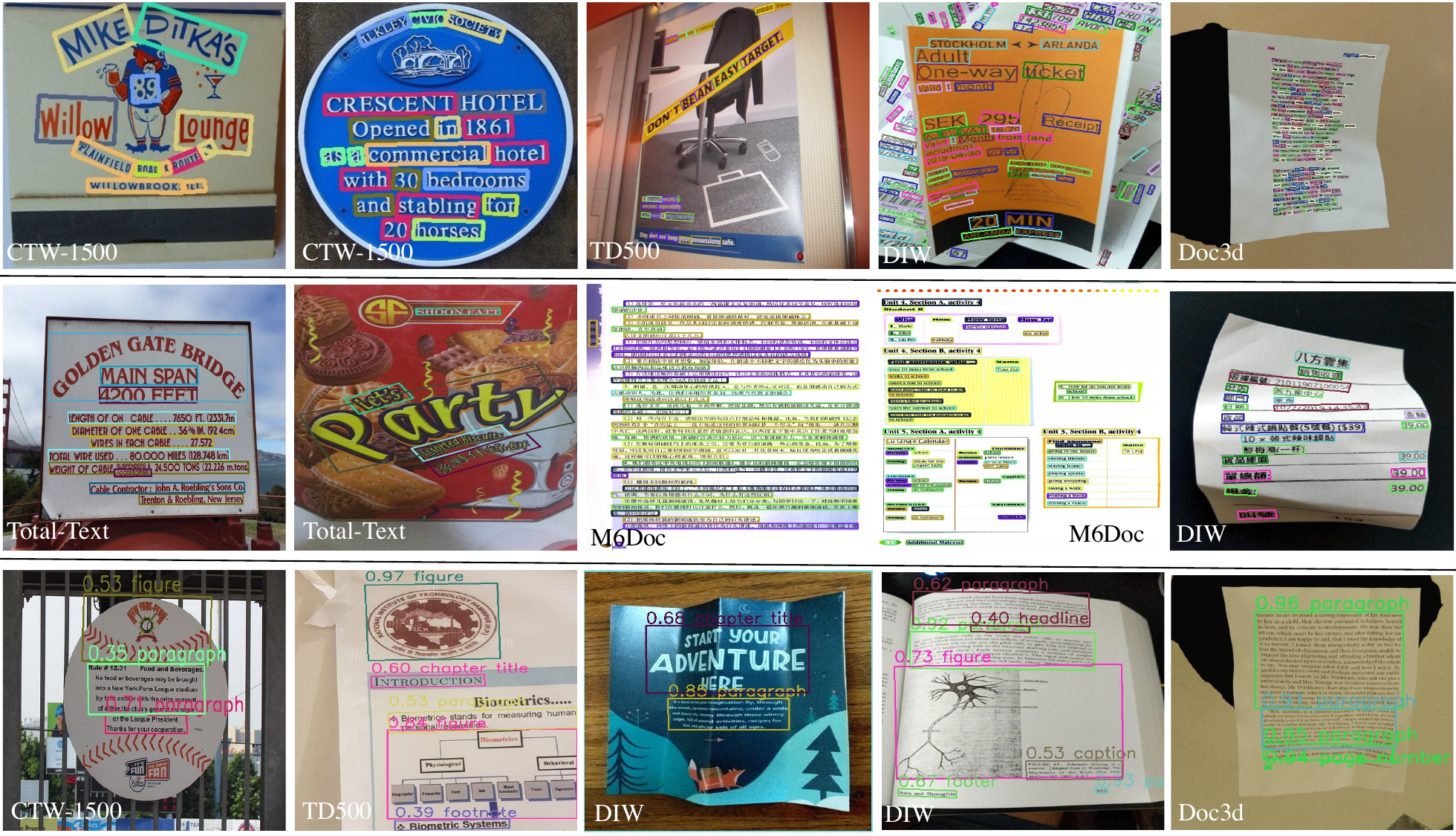}}
\caption{Qualitative results of multi-granularity pseudo labels on public benchmarks produced by our DAT-DET model.}
\label{app_fig2}
\end{center}
%\vskip -0.2in
\end{figure*}

\begin{figure*}[ht]
%\vskip 0.2in
\begin{center}
\centerline{\includegraphics[width=\textwidth]{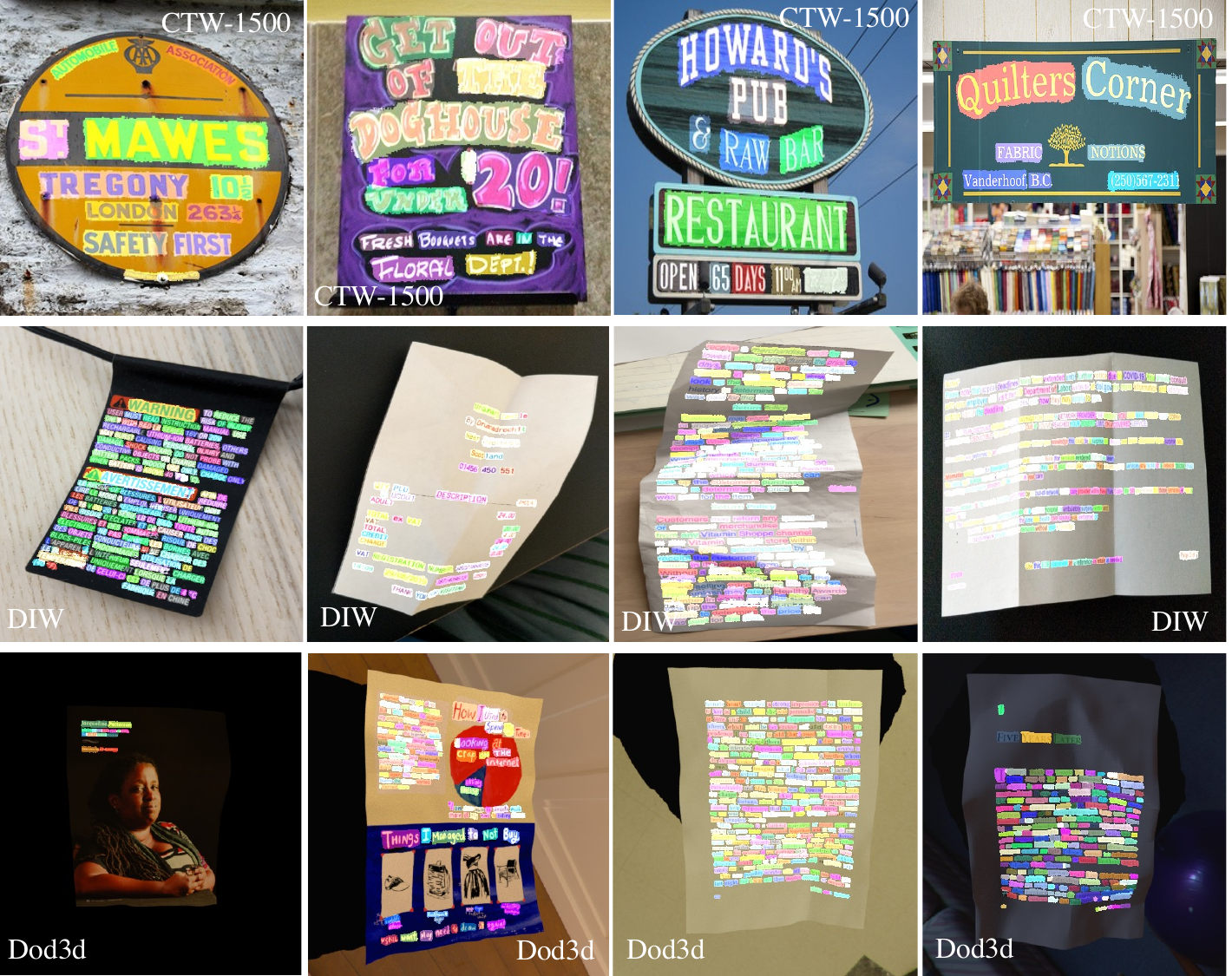}}
\caption{Qualitative results of pseudo labels at word-level granularity on public benchmarks produced by our DAT-SEG model.}
\label{app_fig3}
\end{center}
%\vskip -0.2in
\end{figure*}

\begin{figure*}[ht]
%\vskip 0.2in
\begin{center}
\centerline{\includegraphics[width=\textwidth]{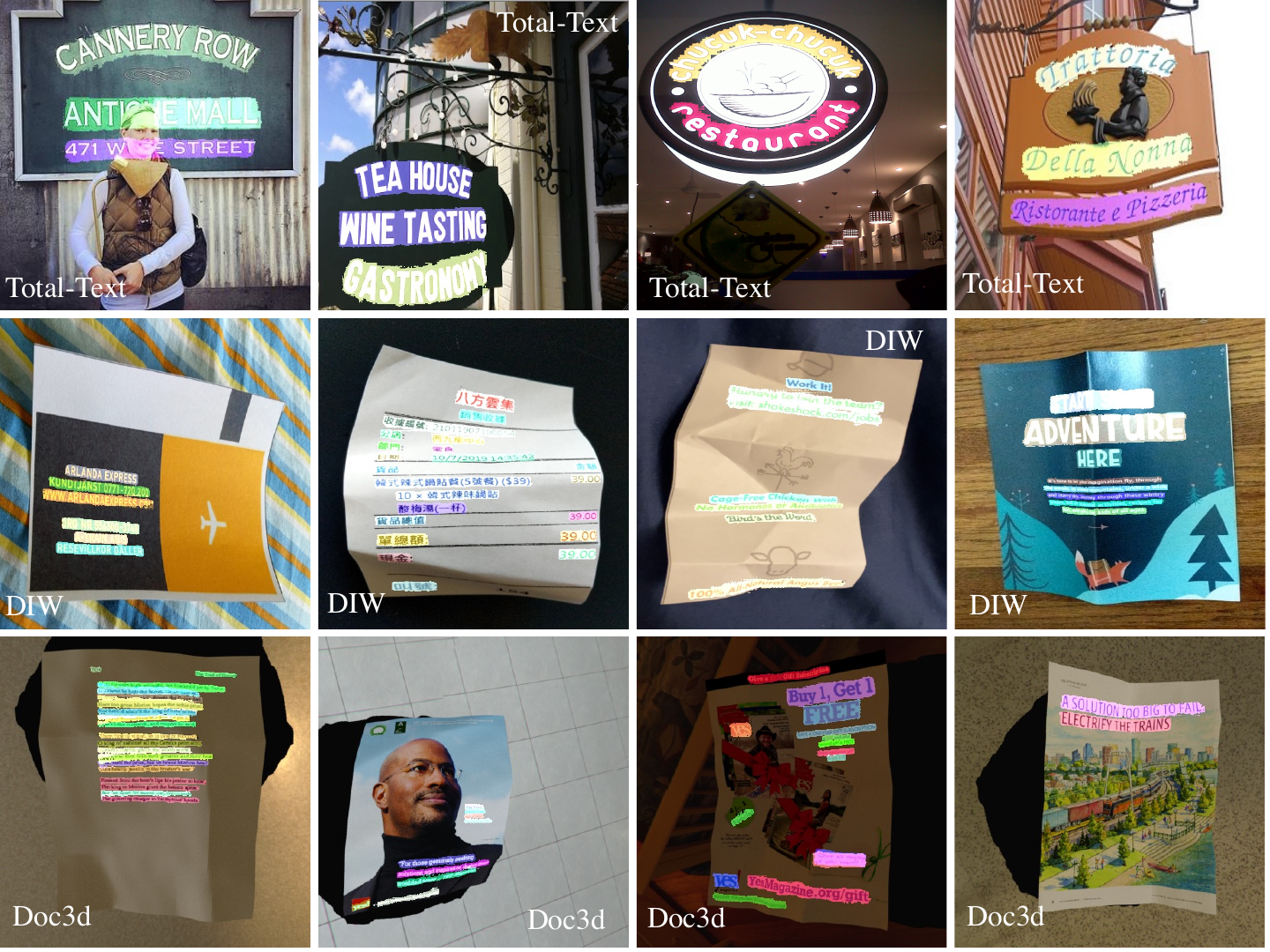}}
\caption{Qualitative results of pseudo labels at text-line level on public benchmarks produced by our DAT-SEG model.}
\label{app_fig4}
\end{center}
%\vskip -0.2in
\end{figure*}

Our proposed multi-granularity text detection framework, equipped with a mixed-granularity training strategy, supports parallel training using datasets with incomplete-granularity annotations. More importantly, after training on multi-granularity public datasets, the resulting DAT model is capable of generating high-quality pseudo labels for various text granularities. This feature significantly enhances the model's utility and applicability, especially in scenarios where comprehensive annotations are not readily available. We provided some qualitative results of multi-granularity pseduo labels produced by our DAT-DET model as shown in~\cref{app_fig2}. It is worth mentioning that the Total-Text dataset is annotated at the word granularity, while the CTW-1500 and MSRA-TD500 datasets are annotated at the line granularity. Additionally, the M6Doc dataset is annotated at the paragraph granularity, the DIW and Doc3D datasets are annotated at the page granularity. \cref{app_fig2} illustrates pseudo labels of text detection branch at different text granularities: the first row shows word-level pseudo labels, the second row presents line-level pseudo labels, and the third row features paragraph-level pseudo labels. \cref{app_fig3} and \cref{app_fig4} further visualize the pseudo labels of segmentation task produced by our DAT-SEG model at the word and line level respectively. Thanks to our well-designed multi-granularity detection framework and interactive cross-granularity representation learning, our model is capable of producing quite promising text detection results at the word and line-level without the need for corresponding annotated data for training. It also demonstrates strong multi-granularity text detection and segmentation capabilities in complex scenarios, such as dense text lines (M6Doc) and rich texts in natural scenes (DIW \& Doc3D).

Our failure cases primarily focus on blurred small text instances and extremely severe occlusions, as shown in the partial image regions at the last row of \cref{app_fig4} in our paper.

%%%%%%%%%%%%%%%%%%%%%%%%%%%%%%%%%%%%%%%%%%%%%%%%%%%%%%%%%%%%%%%%%%%%%%%%%%%%%%%
%%%%%%%%%%%%%%%%%%%%%%%%%%%%%%%%%%%%%%%%%%%%%%%%%%%%%%%%%%%%%%%%%%%%%%%%%%%%%%%
\end{document}